\documentclass[runningheads]{llncs}
\usepackage{graphicx}
\usepackage{amsmath,amssymb}
\usepackage{booktabs}
\usepackage{multirow}
\usepackage{array}
\newcommand{\best}[1]{\textbf{#1}}
\newcommand{\second}[1]{\underline{#1}}
\begin{document}
\title{ReConFuse: Reconstruction-Error Guided
Semantic Fusion for AI-Generated Video
Detection}
\titlerunning{ReConFuse for AI-Generated Video Detection}
\author{
Xiaojing Chen\inst{1} \and
Xinyu Lu\inst{1} \and
Changtao Miao\inst{2} \and
Yunfeng Diao\inst{3}\thanks{Corresponding author: diaoyunfeng@hfut.edu.cn}
}
\authorrunning{X. Chen et al.}

\institute{
Anhui University
\email{chenxiaojing0909@ahu.edu.cn, 2510382668@qq.com}
\and
Ant Group
\email{miaoct@mail.ustc.edu.cn}
\and
Hefei University of Technology
\email{diaoyunfeng@hfut.edu.cn}
}
\maketitle

\begin{abstract}
AI-generated videos are becoming increasingly realistic, raising serious concerns about misinformation, content authenticity, and media trust. Reliable AI-generated video detection is therefore essential for multimedia forensics, yet remains challenging due to the need to capture spatial artifacts, temporal dynamics, and generalize to evolving generative models. In this paper, we explore reconstruction error as a discriminative forensic cue for AI-generated video detection. By reconstructing input videos with a pretrained WF-VAE, we observe that real and generated videos exhibit distinguishable frame-wise reconstruction error patterns, suggesting that reconstruction errors can reveal their distributional discrepancies. However, extending reconstruction-based image detection to videos is non-trivial, since video reconstruction errors are temporally organized across frames and require semantic context for effective interpretation. To address these challenges, we propose \textbf{ReConFuse}, a reconstruction-guided semantic fusion framework for video-level AI-generated video detection. ReConFuse extracts reconstruction error cues from WF-VAE reconstructed videos, aligns them with multi-frame semantic features, and uses a Mamba-based module to model temporal evolution for video-level classification. Experiments across multiple generators and evaluation settings demonstrate the effectiveness and strong generalization ability of ReConFuse.
\vspace{-6mm}
\end{abstract}
\vspace{-4mm}
\section{Introduction}
\vspace{-2mm}
In recent years, advanced video generation models have substantially improved realism and accessibility of AI-generated video. Although these models offer broad potential for content creation, they have also raised persistent concerns regarding misinformation, content authenticity, and media trust~\cite{zhu2025faknow}. Consequently, reliable AI-generated video detection has become an important problem in multimedia forensics~\cite{chang2024what,chen2024demamba,ni2025genvidbench,wen2025busterx,dai2024dmdroid}. At the same time, this task is challenging because it must capture spatial artifacts, model temporal dynamics, and remain robust to rapidly evolving generative models~\cite{gao2025pacl,pang2024mmaf}.

Existing methods for AI-generated video detection primarily exploit visual artifacts, temporal inconsistencies, physical cues, or high-level semantic representations~\cite{ojha2023universal,chen2024demamba,song2024mmdet,zheng2025d3,zhang2025nsgvd}. Different from these approaches, reconstruction-error based cues reveal distributional traces left by generative models from a reconstruction perspective. In contrast, reconstruction-error based cues remain less explored in AI-generated video detection, despite their demonstrated effectiveness in image-level generated content detection~\cite{wang2023dire,ricker2024aeroblade,luo2024lare2,chu2025fire}. This motivates us to investigate whether reconstruction-error cues can also provide useful forensic evidence for AI-generated video detection.

\begin{figure}[t]
    \centering
    \includegraphics[width=0.95\linewidth]{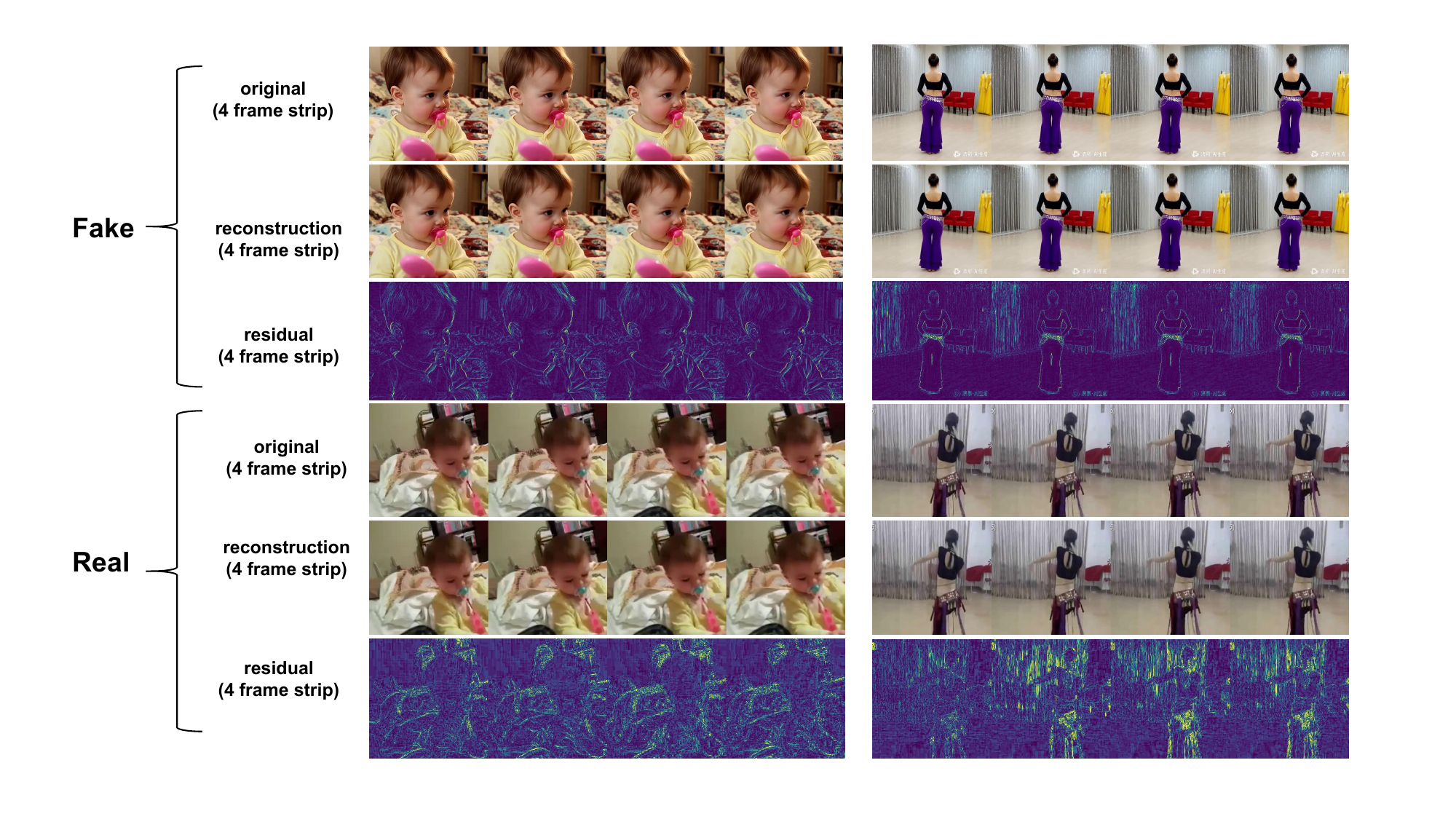}
    \vspace{-2mm}
    \caption{Reconstruction behavior comparison between real and AI-generated videos (generated by Zhipu Qingyan~\cite{zhipu2026qingyan}). The original frames, reconstructed frames  (reconstructed by WF-VAE), and pseudo-color map (Purple denotes low reconstruction error, whereas yellow denotes high reconstruction error). of reconstruction errors illustrate that reconstruction error can provide discriminative low-level forensic evidence for AI-generated video detection.}
    \label{fig:reconstruction_error_observation}
    \vspace{-6mm}
\end{figure}

As an initial investigation, we reconstruct input videos and compute frame-wise reconstruction errors. As shown in Fig.~\ref{fig:reconstruction_error_observation}, real and AI-generated videos exhibit distinguishable error patterns under this pretrained reconstruction prior. However, directly extending image-level reconstruction error analysis to videos is nontrivial: First, unlike image-level reconstruction errors, video reconstruction errors are temporally organized across frames, making independent frame-level modeling insufficient~\cite{bertasius2021timesformer,liu2022videoswin,gu2024mamba}. Second, reconstruction errors evolve over time~\cite{bertasius2021timesformer,liu2022videoswin,gu2024mamba} and can be ambiguous without semantic context~\cite{radford2021clip,ni2022xclip}.

To address these challenges, we propose \textbf{ReConFuse}, a framework that fuses reconstruction-guided error cues with semantic representations for AI-generated video detection. ReConFuse first reconstructs an input video using pretrained WF-VAE~\cite{li2025wfvae} and obtains frame-wise reconstruction error. It then spatially aligns error cues with multi-frame semantic features and uses a Mamba-based sequence module to model their temporal evolution for video-level classification. By integrating low-level reconstruction discrepancies with high-level semantic guidance, ReConFuse aims to improve the reliability and generalization of AI-generated video detection.
The main contributions of this paper are summarized as follows.
\begin{itemize}
\item We investigate the effectiveness of reconstruction error as a discriminative forensic cue for AI-generated video detection and show that real and generated videos exhibit different error patterns under a pretrained video reconstruction prior.
\item We propose ReConFuse, a reconstruction-guided semantic fusion framework that aligns video reconstruction error with semantic features and models their temporal evolution for video-level detection.
\item We conduct extensive experiments across multiple video generation models and evaluation settings, validating the effectiveness and generalization capability of the proposed method.
\end{itemize}
\vspace{-4mm}
\section{Related Work}
\vspace{-2mm}
\textbf{AI-Generated Video Detection}. Early video forgery detectors mainly focused on face-centric manipulations such as DeepFake, FaceSwap, and Face2Face, where facial priors and boundary inconsistencies provide strong cues~\cite{rossler2019faceforensicspp,li2020celebdf,dolhansky2020dfdc,afchar2018mesonet,li2020facexray}. Recent text-to-video and image-to-video generators produce more diverse scenes, objects, and motions, shifting the problem toward general video authenticity analysis~\cite{openai2024sora,blattmann2023svd,singer2022makeavideo,ho2022imagenvideo,villegas2022phenaki}. Existing detectors model this problem from complementary perspectives: DeMamba~\cite{chen2024demamba} models local spatial-temporal details, D3~\cite{zheng2025d3} uses second-order temporal dynamics, MM-Det~\cite{song2024mmdet} combines multimodal and reconstruction-related features, and NSG-VD~\cite{zhang2025nsgvd} analyzes physical consistency. However, reconstruction errors produced by pretrained video generative autoencoders remain underexplored as time-varying forensic cues for AI-generated video detection.

\noindent\textbf{Reconstruction-Error Based Detection}. Reconstruction-error based detection methods use the difference between an input sample and its reconstructed counterpart as forensic evidence~\cite{wang2023dire,ricker2024aeroblade,luo2024lare2,chu2025fire}. Existing reconstruction-error based detection methods have been applied mainly to image forgery detection. DIRE~\cite{wang2023dire} uses diffusion reconstruction error as a discriminative representation for diffusion-generated image detection, while AEROBLADE~\cite{ricker2024aeroblade} shows that the autoencoder component of latent diffusion models can also reveal useful reconstruction discrepancies without running the full diffusion process. LaRE$^2$~\cite{luo2024lare2} further exploits latent reconstruction error for diffusion-generated image detection, and FIRE~\cite{chu2025fire} improves robustness by modeling frequency-guided reconstruction errors. These methods show that reconstruction errors can reveal low-level generation traces.

However, applying image-level reconstruction-error analysis to videos in a frame-wise manner is insufficient for reliable video-level detection because it fails to model how reconstruction-error cues evolve across frames. Simple frame-level aggregation may miss temporally consistent forensic cues across frames. Moreover, reconstruction error primarily reflects low-level visual differences and can be ambiguous in complex scenes without high-level visual representations. To address these limitations, we extend reconstruction-error-based detection to AI-generated video detection by jointly modeling dense frame-wise reconstruction errors, semantic context, and temporal dynamics.
\vspace{-2mm}
\section{Method}
\vspace{-2mm}
This section presents the proposed \textbf{ReConFuse} framework for AI-generated video detection. We first introduce the pretrained priors used in our framework, including WF-VAE for video reconstruction and XCLIP for semantic representation. We then discuss the motivation behind reconstruction-guided semantic-error fusion, and finally describe the full pipeline, including reconstruction error extraction, semantic alignment, gated fusion, and temporal modeling.
\vspace{-2mm}
\subsection{Preliminaries}
\textbf{Latent Diffusion Model and Wavelet-Driven Energy Flow VAE (WF-VAE).}
Latent diffusion models first map images or videos into a latent space through an autoencoder, and then learn the diffusion and reverse denoising processes in that latent space~\cite{rombach2022ldm,blattmann2023svd}. Given a video clip $V=\{x_t\}_{t=1}^{T}$, where $x_t \in \mathbb{R}^{H \times W \times 3}$ denotes the $t$-th frame, a video VAE first encodes it into a spatio-temporal latent representation:
\begin{equation}
z_0 = \mathcal{E}_{v}(V),
\end{equation}
where $\mathcal{E}_{v}$ denotes the video encoder and $z_0$ is the clean latent variable encoded from the input video. To avoid confusion with the frame index $t$, we use $k$ to denote the diffusion step. The forward diffusion process in the latent space can be formulated as:
\begin{equation}
q(z_k|z_0)=\mathcal{N}(z_k;\sqrt{\bar{\alpha}_k}z_0,(1-\bar{\alpha}_k)I),
\end{equation}
where $z_k$ denotes the noisy latent variable at diffusion step $k$, and $\bar{\alpha}_k$ is the predefined noise schedule. Through reverse diffusion sampling, a denoising network gradually recovers a generated latent variable $\tilde{z}_0$, which is then mapped back to the pixel space by the video decoder:
\begin{equation}
\tilde{V}=\mathcal{D}_{v}(\tilde{z}_0),
\end{equation}
where $\mathcal{D}_{v}$ denotes the video decoder and $\tilde{V}$ is the generated video. Therefore, the video VAE serves as the key bridge between the pixel space and the latent space in video latent diffusion pipelines, and its encoder-decoder structure learns the compression and reconstruction regularities of video content.

In this work, we do not perform latent diffusion sampling or diffusion inversion. Inspired by reconstruction-based forensic analysis such as DIRE, we only use the encoder-decoder reconstruction path of a pretrained WF-VAE:
\begin{equation}
z_0=\mathcal{E}_{v}(V), \quad \hat{V}=\mathcal{D}_{v}(z_0).
\end{equation}
This process can be viewed as projecting an input video into the learned video latent reconstruction space and reconstructing it back to the pixel space. We regard the pretrained WF-VAE as an external video reconstruction prior. Since real and AI-generated videos may differ in how well they fit this reconstruction prior, their reconstruction discrepancies can provide useful low-level forensic cues for AI-generated video detection.
\noindent\textbf{XCLIP Semantic Representation.}
XCLIP is a pretrained vision-language model that provides semantically rich video representations~\cite{qin2025dvlp}. In this work, we use only its visual encoder to extract semantic features from input frames, rather than using its text retrieval branch or similarity aggregation module. For each frame $x_t$, the XCLIP visual encoder produces patch-level semantic tokens:
\begin{equation}
S_t = \mathrm{XCLIP}_{\mathrm{vis}}(x_t), \quad S_t=\{s_t^{(i)}\}_{i=1}^{N},
\end{equation}
where $s_t^{(i)}$ denotes the semantic token of the $i$-th spatial patch and $N$ is the number of patches. These tokens provide semantic context for the corresponding reconstruction-error tokens, helping the model distinguish normal content-dependent reconstruction difficulties from generation-related discrepancies.
\vspace{-2mm}
\subsection{Reconstruction-Guided Semantic-Error Fusion}
\begin{figure}[t]
    \centering
    \includegraphics[width=\linewidth]{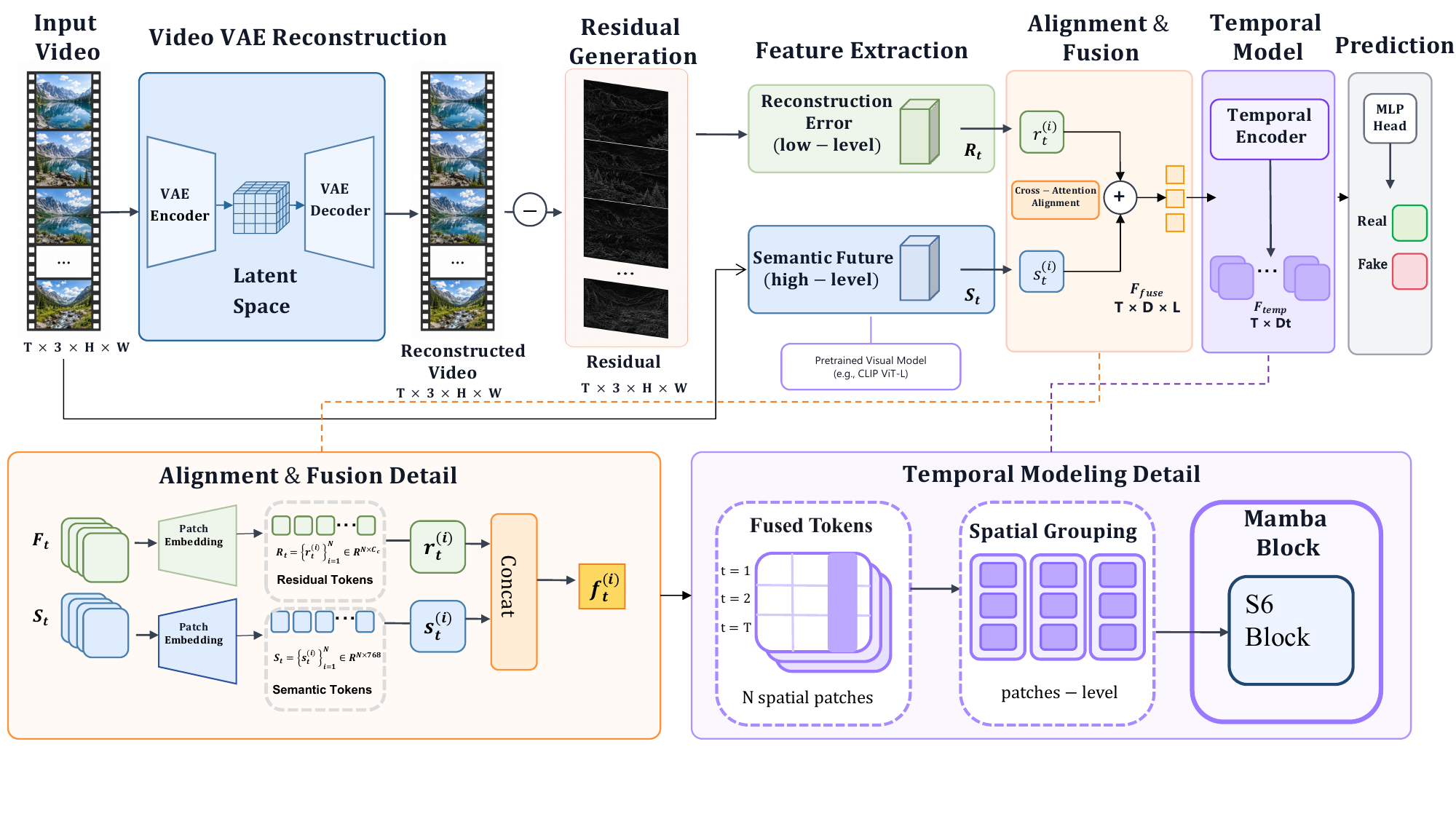}
    \vspace{-8mm}
    \caption{Overview of the proposed ReConFuse framework. The input video is reconstructed by a pretrained video VAE, reconstruction errors are extracted as low-level forensic evidence, reconstruction error tokens are aligned and fused with semantic tokens, and Mamba is used for temporal modeling and video-level real/fake prediction.}
    \label{fig:method_overview}
    \vspace{-6mm}
\end{figure}

\vspace{-2mm}
As shown in Fig.~\ref{fig:method_overview}, ReConFuse preserves dense frame-wise reconstruction error maps, aligns them with XCLIP semantic tokens at the patch level, and aggregates the fused representation for video-level real/fake prediction.
\vspace{-5mm}
\subsubsection{Reconstruction Error Representation.}
Given the reconstructed video $\hat{V}=\{\hat{x}_t\}_{t=1}^{T}$ produced by WF-VAE, we compute the signed reconstruction error for each frame:
\begin{equation}
e_t = x_t - \hat{x}_t,
\end{equation}
where $x_t$ and $\hat{x}_t$ denote the original and reconstructed frames at time $t$, respectively. We use signed reconstruction errors rather than magnitude-only errors, since the direction of reconstruction discrepancy may contain useful forensic information. The reconstruction error sequence is denoted as:
\begin{equation}
E=\{e_t\}_{t=1}^{T}.
\end{equation}
To preserve local forensic cues, we convert each reconstruction error map into patch-level reconstruction error tokens. Specifically, each reconstruction error map is resized to the input resolution of the XCLIP visual encoder and embedded using a lightweight patch embedding layer:
\begin{equation}
P_t = \phi_p\left(\mathcal{R}(e_t)\right),
\end{equation}
where:
\begin{equation}
P_t=\{p_t^{(i)}\}_{i=1}^{N}, \quad P_t \in \mathbb{R}^{N \times C_p}.
\end{equation}
Here, $\mathcal{R}(\cdot)$ denotes the resizing operation that matches the input resolution of the XCLIP visual encoder, $\phi_p(\cdot)$ denotes a lightweight patch embedding layer, $p_t^{(i)}$ denotes the reconstruction error token of the $i$-th patch, $N$ is the number of patches, and $C_p$ is the reconstruction error token dimension. Compared with scalar reconstruction error, this dense reconstruction error representation preserves localized reconstruction anomalies such as texture distortions, boundary inconsistencies, and structural deviations.
\vspace{-5mm}
\subsubsection{Semantic-Error Fusion.}
Reconstruction errors alone can be ambiguous because large errors may also arise from motion blur, compression artifacts, illumination changes, or complex textures. Therefore, we fuse reconstruction error tokens with XCLIP semantic tokens to interpret reconstruction discrepancies under semantic context.

For each frame $x_t$, the pretrained XCLIP visual encoder extracts patch-level semantic tokens:
\begin{equation}
S_t=\{s_t^{(i)}\}_{i=1}^{N}, \quad S_t \in \mathbb{R}^{N \times C_s}.
\end{equation}
Since reconstruction error and semantic tokens are generated on the same patch grid, $p_t^{(i)}$ and $s_t^{(i)}$ correspond to the same local region. We first project them into a shared feature dimension:
\begin{equation}
\tilde{s}_t^{(i)} = W_s s_t^{(i)}, \quad \tilde{p}_t^{(i)} = W_p p_t^{(i)}.
\end{equation}
Then, a reconstruction-error-guided gate is used to estimate how much reconstruction error evidence should be injected into the semantic representation:
\begin{equation}
a_t^{(i)} = \sigma\left(W_a[\tilde{s}_t^{(i)};\tilde{p}_t^{(i)}]\right),
\end{equation}
where $[\cdot;\cdot]$ denotes channel-wise concatenation and $\sigma(\cdot)$ is the sigmoid function. The fused semantic-error token is computed as:
\begin{equation}
f_t^{(i)} = \tilde{s}_t^{(i)} + a_t^{(i)} \odot \tilde{p}_t^{(i)},
\end{equation}
where $\odot$ denotes element-wise multiplication. This gated fusion enables the model to emphasize reconstruction error cues that are informative under the corresponding semantic context, while suppressing reconstruction error variations caused by normal video content. The fused semantic-error token sequence of frame $t$ is:
\begin{equation}
F_t=\{f_t^{(i)}\}_{i=1}^{N}, \quad F_t \in \mathbb{R}^{N \times C_f}.
\end{equation}
By stacking all frames, we obtain:
\begin{equation}
F=\{F_t\}_{t=1}^{T}.
\end{equation}
\subsubsection{Video-Level Prediction.}
The fused semantic-error representation $F$ is fed into a Mamba-based temporal module to aggregate cross-frame forensic evidence. The temporal module captures how local reconstruction anomalies evolve across frames and outputs a video-level representation $z$. To preserve global semantic information, we concatenate $z$ with the video-level semantic context $g$:
\begin{equation}
h=[z;g].
\end{equation}
Finally, the probability that the input video is AI-generated is predicted by:
\begin{equation}
\hat{y}=\sigma(Wh+b).
\end{equation}
The model is trained with the binary cross-entropy loss:
\begin{equation}
\mathcal{L}=-\left[y\log(\hat{y})+(1-y)\log(1-\hat{y})\right],
\end{equation}
where $y \in \{0,1\}$ is the ground-truth label.
\vspace{-2mm}
\section{Experiments}
\vspace{-2mm}
In this section, we evaluate the proposed ReConFuse framework on two AI-generated video detection benchmarks, GenVideo and GenBuster. The experiments are designed to assess detection performance under the one-to-many and many-to-many settings of GenVideo, as well as the benchmark protocol of GenBuster. We compare ReConFuse with representative video-level and image-level detectors, and further conduct ablation studies to analyze the contribution of reconstruction error cues, semantic alignment, and temporal modeling.

\textbf{Datasets.}
We evaluate our method on two AI-generated video detection benchmarks: GenVideo~\cite{chen2024demamba} and GenBuster~\cite{wen2025busterx}. GenVideo is a large-scale benchmark introduced by DeMamba, containing real-world videos and AI-generated videos from diverse generative models. Following the standard GenVideo protocol, we evaluate on both one-to-many and many-to-many settings. In the one-to-many setting, models are trained with 10,000 real videos from Kinetics-400~\cite{kay2017kinetics} and 10,000 AI-generated videos from Pika~\cite{pang2024vgmshield}. In the many-to-many generalization task, models are trained on 10 baseline categories and evaluated the average detection performance on MSR-VTT~\cite{xu2016msrvtt} and 10 diverse
AI-generated datasets from different generation paradigms. For GenBuster, we follow its benchmark protocol and evaluate on subsets generated by recent commercial 8 generators.

\textbf{Evaluation Metrics.}
We evaluate performance using Recall, Accuracy, F1-score~\cite{chinchor1993muc5}, and AUROC~\cite{huang2005auc}, treating AI-generated videos as the positive class. All results are percentages; bold and underlined values indicate the best and second-best results.

\textbf{Baselines.}
We compare ReConFuse with representative baselines, including DeMamba~\cite{chen2024demamba}, DIRE~\cite{wang2023dire}, AEROBLADE~\cite{ricker2024aeroblade}, and D3~\cite{zheng2025d3}. DeMamba, D3, and ReConFuse are video-level detectors, while DIRE and AEROBLADE are image-level detectors evaluated by aggregating frame-level predictions. DeMamba, DIRE, and ReConFuse require training, whereas D3 and AEROBLADE are training-free methods. For training-based methods, we follow the same training split whenever applicable. Training-free methods are included as reference forensic detectors and are evaluated without benchmark-specific training.
\vspace{-4mm}
\subsection{Main Results on GenVideo}
\vspace{-2mm}
\subsubsection{One-to-Many Setting.}
Table~\ref{tab:genvideo_one_to_many} reports the results on GenVideo under the one-to-many setting. Our method achieves the best average Accuracy and F1-score, improving over DeMamba from 79.56\% to 85.56\% Accuracy and from 71.75\% to 85.03\% F1-score.

D3 obtains the highest average Recall, and DIRE obtains the highest average AUROC, but our method provides stronger Accuracy and F1-score under the fixed decision threshold.
\vspace{-5mm}
\begin{table}[t]
\centering
\caption{GenVideo one-to-many results. Values are percentages. Best and second-best results are bolded and underlined.}
\label{tab:genvideo_one_to_many}
\begingroup
\setlength{\tabcolsep}{1.2pt}
\renewcommand{\arraystretch}{1.08}
\resizebox{\linewidth}{!}{
\begin{tabular}{l c c c *{11}{c}}
\toprule
\textbf{Method} & \textbf{Mod.} & \textbf{Train} & \textbf{Metric} & \textbf{Scope} & \textbf{Morph} & \textbf{Moon} & \textbf{Hot.} & \textbf{Show1} & \textbf{Gen2} & \textbf{Craft.} & \textbf{Lavie} & \textbf{Sora} & \textbf{Wild} & \textbf{Avg.} \\
\midrule
\multirow{4}{*}{DeMamba} & \multirow{4}{*}{Video} & \multirow{4}{*}{Yes} & Recall & 74.20 & 81.40 & 94.96 & 28.00 & 30.60 & 81.20 & 72.80 & 39.60 & 42.86 & 50.55 & 59.62 \\
& & & Acc. & \second{86.70} & \best{90.30} & \second{97.13} & 63.60 & 64.90 & \second{90.20} & 86.00 & 69.40 & \second{71.43} & \second{75.97} & 79.56 \\
& & & F1 & \second{84.80} & \second{89.35} & \second{97.00} & 43.48 & 46.58 & 89.23 & 83.87 & 56.41 & 60.00 & 66.76 & 71.75 \\
& & & AUROC & \best{96.31} & \best{98.47} & \best{99.87} & \best{85.34} & 85.17 & \second{99.02} & \best{98.13} & \best{90.23} & 89.03 & 86.23 & \second{92.78} \\
\midrule
\multirow{4}{*}{DIRE} & \multirow{4}{*}{Image} & \multirow{4}{*}{Yes} & Recall & 61.20 & 80.00 & \second{98.00} & 16.00 & 33.00 & 91.20 & 80.60 & 34.60 & 35.71 & 43.20 & 57.35 \\
& & & Acc. & 79.80 & 89.20 & \best{98.20} & 57.20 & 65.70 & \best{94.80} & \second{89.50} & 66.50 & 67.86 & 70.80 & 77.96 \\
& & & F1 & 75.18 & 88.11 & \best{98.20} & 27.21 & 49.03 & \best{94.61} & 88.47 & 50.81 & 52.63 & 59.67 & 68.39 \\
& & & AUROC & 93.05 & 97.18 & \second{99.66} & 82.97 & \second{90.50} & \best{99.13} & \second{97.87} & 87.54 & \second{90.47} & \best{91.84} & \best{93.02} \\
\midrule
\multirow{4}{*}{AEROBLADE} & \multirow{4}{*}{Image} & \multirow{4}{*}{No} & Recall & 58.60 & 75.00 & 79.40 & 60.20 & 62.00 & 77.80 & 88.20 & 43.80 & 33.93 & 35.80 & 61.47 \\
& & & Acc. & 78.80 & 87.00 & 89.20 & \best{79.60} & \second{80.50} & 88.40 & \best{93.60} & \second{71.40} & 66.07 & 67.40 & \second{80.20} \\
& & & F1 & 73.43 & 85.23 & 88.03 & \best{74.69} & \second{76.07} & 87.02 & \best{93.23} & 60.50 & 50.00 & 52.34 & \second{74.05} \\
& & & AUROC & -- & -- & -- & -- & -- & -- & -- & -- & -- & -- & -- \\
\midrule
\multirow{4}{*}{D3} & \multirow{4}{*}{Video} & \multirow{4}{*}{No} & Recall & \best{92.50} & \second{88.00} & 94.00 & \best{84.50} & \best{83.00} & \best{96.50} & \best{93.00} & \best{85.00} & \best{99.00} & \best{87.50} & \best{90.30} \\
& & & Acc. & 64.50 & 69.00 & 66.00 & 61.50 & 60.00 & 56.50 & 63.50 & 68.50 & 51.25 & 59.50 & 62.03 \\
& & & F1 & 71.80 & 73.95 & 73.60 & 69.20 & 68.10 & 68.80 & 71.60 & \second{72.96} & \second{67.01} & \second{68.90} & 70.59 \\
& & & AUROC & -- & -- & -- & -- & -- & -- & -- & -- & -- & -- & -- \\
\midrule
\multirow{4}{*}{Ours} & \multirow{4}{*}{Video} & \multirow{4}{*}{Yes} & Recall & \second{89.20} & \best{94.00} & \best{99.79} & \second{65.20} & \second{80.80} & \second{94.00} & \second{91.40} & \second{72.00} & \second{83.93} & \second{77.46} & \second{84.78} \\
& & & Acc. & \best{87.70} & \second{90.10} & 92.83 & \second{75.70} & \best{83.50} & 90.10 & 88.80 & \best{79.10} & \best{85.71} & \best{82.03} & \best{85.56} \\
& & & F1 & \best{87.88} & \best{90.47} & 93.14 & \second{72.85} & \best{83.04} & \second{90.47} & \second{89.08} & \best{77.50} & \best{85.45} & \best{80.45} & \best{85.03} \\
& & & AUROC & \second{95.80} & \second{97.37} & 99.54 & \second{83.73} & \best{91.11} & 95.46 & 95.21 & \second{88.11} & \best{92.70} & \second{88.19} & 92.72 \\
\bottomrule
\end{tabular}}
\endgroup
\vspace{-4mm}
\end{table}
\subsubsection{Many-to-Many Setting.}
Table~\ref{tab:genvideo_many_to_many} shows the comparison results on GenVideo under the more diverse many-to-many setting.

Our method achieves the best average Recall, Accuracy, F1-score, and AUROC, improving over DeMamba from 90.17\% to 91.93\% Accuracy and from 89.52\% to 91.49\% F1-score.

D3 remains recall-oriented, whereas our method achieves more balanced video-level performance, especially in average F1-score.
\begin{table}[t]
\centering
\caption{GenVideo many-to-many results. Values are percentages. Best and second-best results are bolded and underlined.}
\label{tab:genvideo_many_to_many}
\begingroup
\setlength{\tabcolsep}{1.2pt}
\renewcommand{\arraystretch}{1.08}
\resizebox{\linewidth}{!}{
\begin{tabular}{l c c c *{11}{c}}
\toprule
\textbf{Method} & \textbf{Mod.} & \textbf{Train} & \textbf{Metric} & \textbf{Scope} & \textbf{Morph} & \textbf{Moon} & \textbf{Hot.} & \textbf{Show1} & \textbf{Gen2} & \textbf{Craft.} & \textbf{Lavie} & \textbf{Sora} & \textbf{Wild} & \textbf{Avg.} \\
\midrule
\multirow{4}{*}{DeMamba} & \multirow{4}{*}{Video} & \multirow{4}{*}{Yes} & Recall & 69.40 & \second{99.60} & \second{99.79} & \best{96.40} & \second{97.00} & \second{99.60} & \second{99.40} & \best{95.20} & \second{55.36} & \best{94.53} & \second{90.63} \\
& & & Acc. & \second{79.70} & \second{94.80} & 94.77 & \best{93.20} & \second{93.50} & 94.80 & \second{94.70} & \best{92.60} & \second{71.43} & \best{92.16} & \second{90.17} \\
& & & F1 & \second{77.37} & \second{95.04} & 94.91 & \best{93.41} & \second{93.72} & 95.04 & \second{94.94} & \best{92.79} & 65.96 & \best{92.01} & \second{89.52} \\
& & & AUROC & \second{89.72} & \second{99.55} & \second{99.44} & \best{97.81} & \second{98.36} & \second{99.48} & \second{99.31} & \best{96.91} & \second{76.40} & \best{96.45} & \second{95.40} \\
\midrule
\multirow{4}{*}{DIRE} & \multirow{4}{*}{Image} & \multirow{4}{*}{Yes} & Recall & 46.40 & 76.40 & 69.80 & 63.80 & 56.00 & 75.00 & 83.80 & 58.80 & 35.71 & 27.40 & 59.31 \\
& & & Acc. & 71.40 & 86.40 & 83.10 & 80.10 & 76.20 & 85.70 & 90.10 & 77.60 & 66.96 & 61.90 & 77.95 \\
& & & F1 & 61.87 & 84.89 & 80.51 & 76.22 & 70.18 & 83.99 & 89.43 & 72.41 & 51.95 & 41.83 & 71.33 \\
& & & AUROC & 85.73 & 96.01 & 93.79 & 91.44 & 89.96 & 95.13 & 96.87 & 89.46 & \best{84.15} & 76.66 & 89.92 \\
\midrule
\multirow{4}{*}{AEROBLADE} & \multirow{4}{*}{Image} & \multirow{4}{*}{No} & Recall & 51.20 & 65.20 & 93.40 & 32.00 & 61.60 & 94.80 & 81.80 & 49.20 & 25.00 & 53.60 & 60.78 \\
& & & Acc. & 75.10 & 82.10 & \best{96.20} & 65.50 & 80.30 & \best{96.90} & 90.40 & 74.10 & 61.61 & 76.30 & 79.85 \\
& & & F1 & 67.28 & 78.46 & \best{96.09} & 48.12 & 75.77 & \best{96.83} & 89.50 & 65.51 & 39.44 & 69.34 & 72.63 \\
& & & AUROC & -- & -- & -- & -- & -- & -- & -- & -- & -- & -- & -- \\
\midrule
\multirow{4}{*}{D3} & \multirow{4}{*}{Video} & \multirow{4}{*}{No} & Recall & \second{92.50} & 88.00 & 94.00 & 84.50 & 83.00 & 96.50 & 93.00 & 85.00 & \best{99.00} & 87.50 & 90.30 \\
& & & Acc. & 64.50 & 69.00 & 66.00 & 61.50 & 60.00 & 56.50 & 63.50 & 68.50 & 51.25 & 59.50 & 62.03 \\
& & & F1 & 71.80 & 73.95 & 73.60 & 69.20 & 68.10 & 68.80 & 71.60 & 72.96 & \best{67.01} & 68.90 & 70.59 \\
& & & AUROC & -- & -- & -- & -- & -- & -- & -- & -- & -- & -- & -- \\
\midrule
\multirow{4}{*}{Ours} & \multirow{4}{*}{Video} & \multirow{4}{*}{Yes} & Recall & \best{97.40} & \best{100.00} & \best{100.00} & \second{95.60} & \best{99.60} & \best{99.80} & \best{100.00} & \second{93.80} & \second{55.36} & \second{92.78} & \best{93.44} \\
& & & Acc. & \best{94.00} & \best{95.30} & \second{95.18} & \second{93.10} & \best{95.10} & \second{95.20} & \best{95.30} & \second{92.20} & \best{72.32} & \second{91.64} & \best{91.93} \\
& & & F1 & \best{94.20} & \best{95.51} & \second{95.30} & \second{93.27} & \best{95.31} & \second{95.41} & \best{95.51} & \second{92.32} & \second{66.67} & \second{91.38} & \best{91.49} \\
& & & AUROC & \best{97.66} & \best{99.84} & \best{99.77} & \second{97.32} & \best{99.07} & \best{99.78} & \best{99.70} & \second{96.30} & 72.67 & \second{96.08} & \best{95.82} \\
\bottomrule
\end{tabular}}
\endgroup
\vspace{-4mm}
\end{table}
\vspace{-4mm}
\subsection{Evaluation on GenBuster}
\vspace{-2mm}
We further evaluate different methods on GenBuster to verify the effectiveness of ReConFuse on an additional AI-generated video detection benchmark. Following the GenBuster benchmark protocol, we report results on subsets generated by recent commercial generators, which provide a complementary evaluation scenario beyond GenVideo. Table~\ref{tab:genbuster} reports the comparison results.
\begin{table}[t]
\centering
\caption{Evaluation on GenBuster under its benchmark protocol. Values are percentages. Best and second-best results are bolded and underlined.}
\label{tab:genbuster}
\begingroup
\setlength{\tabcolsep}{1.8pt}
\renewcommand{\arraystretch}{1.08}
\resizebox{\linewidth}{!}{
\begin{tabular}{l c c c *{9}{c}}
\toprule
\textbf{Method} & \textbf{Mod.} & \textbf{Train} & \textbf{Metric} & \textbf{Gen3} & \textbf{Jimeng} & \textbf{Kling} & \textbf{Luma} & \textbf{Pika} & \textbf{Sora} & \textbf{Vidu} & \textbf{Wanx} & \textbf{Avg.} \\
\midrule
\multirow{4}{*}{DeMamba} & \multirow{4}{*}{Video} & \multirow{4}{*}{Yes} & Recall & 86.00 & 79.00 & \second{83.00} & \second{93.00} & \second{86.00} & 79.50 & 92.67 & \second{96.00} & 86.90 \\
& & & Acc. & \best{96.67} & \second{95.50} & \second{96.17} & \second{97.83} & \second{96.67} & \second{93.29} & \best{97.38} & \best{98.15} & \second{96.46} \\
& & & F1 & \second{89.58} & \second{85.41} & \second{87.83} & \second{93.47} & \second{89.58} & \second{87.12} & \second{94.24} & \best{96.00} & \second{90.40} \\
& & & AUROC & \second{99.38} & \best{98.62} & \second{98.79} & \second{99.50} & \second{99.18} & \best{99.12} & \second{99.65} & \second{99.79} & \second{99.25} \\
\midrule
\multirow{4}{*}{DIRE} & \multirow{4}{*}{Image} & \multirow{4}{*}{Yes} & Recall & 25.80 & 64.80 & 68.40 & 46.20 & 29.20 & 67.20 & 70.00 & 44.40 & 52.00 \\
& & & Acc. & 62.70 & 82.20 & 84.00 & 72.90 & 64.40 & 83.40 & 84.80 & 72.00 & 75.80 \\
& & & F1 & 40.89 & 78.45 & 81.04 & 63.03 & 45.06 & 80.19 & 82.16 & 61.33 & 66.52 \\
& & & AUROC & 85.14 & 95.74 & 96.87 & 89.46 & 83.22 & 96.09 & 96.23 & 90.36 & 91.64 \\
\midrule
\multirow{4}{*}{AEROBLADE} & \multirow{4}{*}{Image} & \multirow{4}{*}{No} & Recall & 26.79 & 25.00 & 66.40 & 69.00 & 26.79 & 73.80 & 70.80 & 43.40 & 50.25 \\
& & & Acc. & \second{63.39} & 62.30 & 83.10 & 84.40 & 63.39 & 86.90 & \second{85.40} & 71.70 & 75.07 \\
& & & F1 & 42.25 & 39.87 & 79.71 & 81.56 & 42.25 & 84.93 & 82.90 & 60.53 & 64.25 \\
& & & AUROC & -- & -- & -- & -- & -- & -- & -- & -- & -- \\
\midrule
\multirow{4}{*}{D3} & \multirow{4}{*}{Video} & \multirow{4}{*}{No} & Recall & \best{100.00} & \best{88.00} & \second{83.00} & 85.00 & 83.00 & \best{99.00} & \best{99.33} & \best{100.00} & \best{92.17} \\
& & & Acc. & 50.00 & 69.00 & 72.50 & 68.50 & 67.50 & 51.25 & 50.67 & 62.50 & 61.49 \\
& & & F1 & 66.67 & 73.95 & 75.11 & 72.96 & 71.86 & 67.01 & 66.82 & 72.73 & 70.89 \\
& & & AUROC & -- & -- & -- & -- & -- & -- & -- & -- & -- \\
\midrule
\multirow{4}{*}{Ours} & \multirow{4}{*}{Video} & \multirow{4}{*}{Yes} & Recall & \second{88.00} & \second{82.00} & \best{89.00} & \best{97.00} & \best{90.00} & \second{82.00} & \second{94.00} & \second{96.00} & \second{89.75} \\
& & & Acc. & \best{96.67} & \best{95.67} & \best{96.83} & \best{98.17} & \best{97.00} & \best{93.71} & \best{97.38} & \second{97.85} & \best{96.66} \\
& & & F1 & \best{89.80} & \best{86.32} & \best{90.36} & \best{94.63} & \best{90.91} & \best{88.17} & \best{94.31} & \second{95.36} & \best{91.23} \\
& & & AUROC & \best{99.61} & \second{98.50} & \best{99.17} & \best{99.64} & \best{99.44} & \second{99.10} & \best{99.70} & \best{99.81} & \best{99.37} \\
\bottomrule
\end{tabular}}
\endgroup
\vspace{-3mm}
\end{table}
On GenBuster, our method achieves the best average Accuracy, F1-score, and AUROC, improving DeMamba from 90.40\% to 91.23\% F1-score and from 99.25\% to 99.37\% AUROC.
\vspace{-7mm}
\subsection{Ablation Study}
\vspace{-2mm}
We conduct ablation studies to analyze the contribution of each major component in ReConFuse. Unless otherwise specified, all ablation results are reported in percentage form using Accuracy, F1-score, and AUROC.
\vspace{-4mm}
\subsubsection{Effect of Major Components.}
Table~\ref{tab:ablation_components} reports the effect of removing or replacing the main components. The complete model improves the XCLIP-only baseline from 79.56\% to 88.33\% Accuracy and from 77.14\% to 88.18\% F1-score, showing the benefit of WF-VAE reconstruction cues.

Error-only detection performs poorly, indicating that reconstruction discrepancies require semantic context. Removing patch-level alignment or replacing Mamba with mean pooling also degrades performance, confirming the importance of local semantic correspondence and temporal modeling.
\vspace{-4mm}
\begin{table}[!htbp]
\centering
\caption{Ablation study on major components. Values are percentages.}
\label{tab:ablation_components}
\begingroup
\setlength{\tabcolsep}{3pt}
\renewcommand{\arraystretch}{1.08}
\scalebox{0.75}{ \begin{tabular}{l c c c c c c c}
\toprule
\textbf{Variant} & \textbf{Error} & \textbf{Sem.} & \textbf{Patch} & \textbf{Temp.} & \textbf{Acc.} & \textbf{F1} & \textbf{AUC} \\
\midrule
XCLIP only & -- & \checkmark & -- & \checkmark & 79.56 & 77.14 & 82.78 \\
Error only & \checkmark & -- & -- & \checkmark & 43.65 & 68.05 & 80.40 \\
w/o Semantic-Error Fusion & \checkmark & \checkmark & -- & \checkmark & 60.11 & 67.79 & 84.28 \\
w/o Mamba & \checkmark & \checkmark & \checkmark & -- & 68.55 & 75.63 & 86.17 \\
Ours & \checkmark & \checkmark & \checkmark & \checkmark & \best{88.33} & \best{88.18} & \best{95.15} \\
\bottomrule
\end{tabular} }
\endgroup
\vspace{-7mm}
\end{table}
\vspace{-4mm}
\subsubsection{Effect of Different VAE Reconstruction Priors.}
We further study the influence of different VAE reconstruction priors for extracting reconstruction-error cues. Specifically, we compare reconstruction errors produced by Wan 2.2 VAE, CogVideoX VAE, and SD1.5 VAE under the same downstream detection framework. This experiment aims to determine whether the choice of reconstruction prior affects the quality of forensic evidence used by ReConFuse.

\begin{figure}[t]
    \centering
    \includegraphics[width=\linewidth]{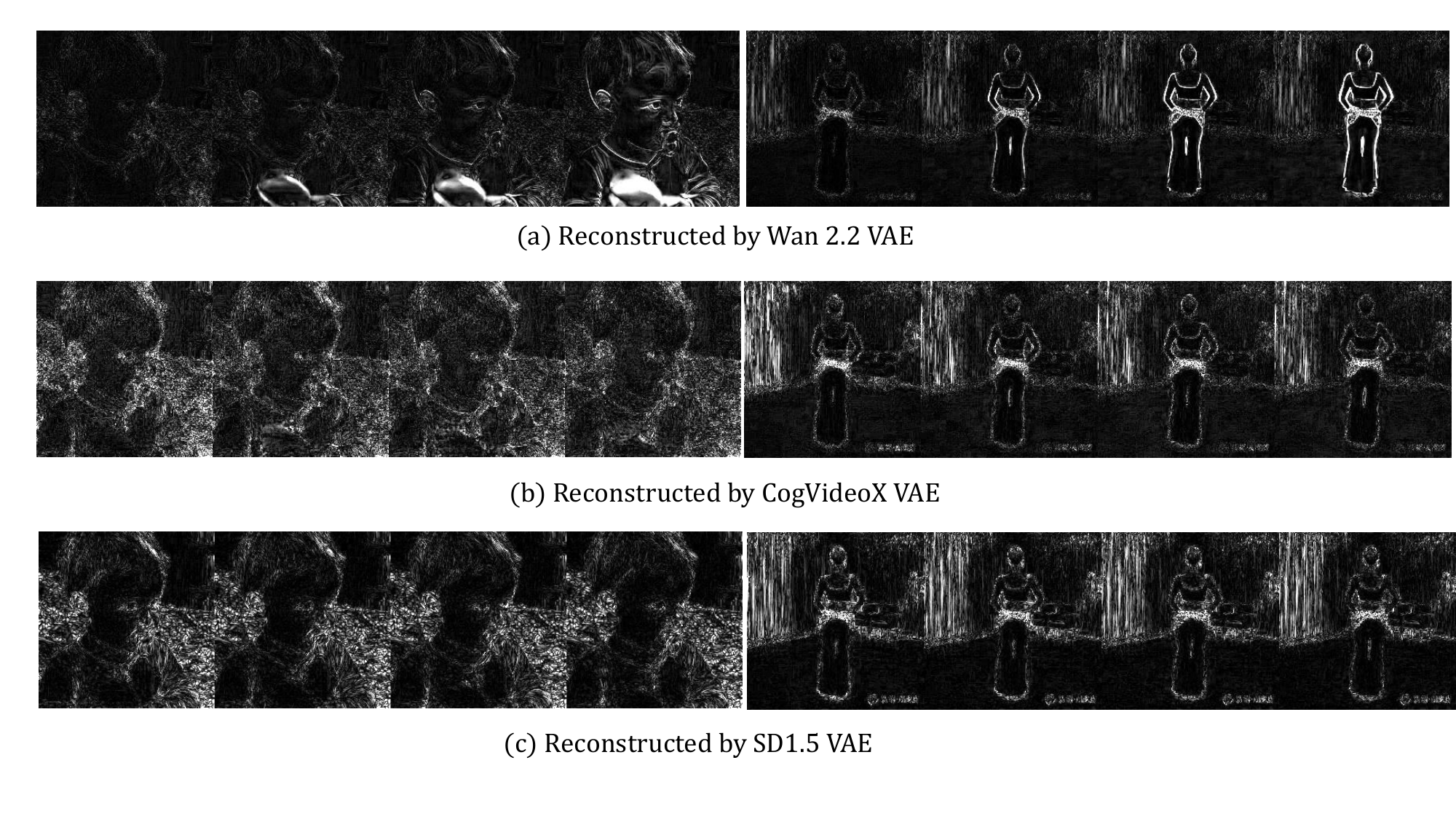}
    \vspace{-4mm}
    \caption{Comparison of grayscale reconstruction-error maps produced by different VAE reconstruction priors. This comparison supports the use of a video reconstruction prior for AI-generated video detection.}
    \label{fig:vae_prior_selection}
    \vspace{-4mm}
\end{figure}

As shown in Fig.~\ref{fig:vae_prior_selection}, different VAEs produce noticeably different reconstruction-error patterns. The video VAEs generate temporally organized error responses across consecutive frames, while the image-level SD1.5 VAE tends to produce more frame-wise texture-dominated residuals.This indicates that, compared with these VAEs, WF-VAE produces smaller reconstruction errors and can better fit the generation process.
Based on this comparison, we adopt WF-VAE as the reconstruction prior in ReConFuse, since it provides dense frame-wise reconstruction discrepancies while maintaining temporal consistency across frames.
\vspace{-5mm}
\subsubsection{Effect of Reconstruction Error Formulation.}
We further compare different WF-VAE reconstruction error VAEs. As shown in Table~\ref{tab:ablation_error}, This suggests that the direction of reconstruction deviation provides useful information that magnitude-only formulations discard.
\vspace{-2mm}
\begin{table}[!htbp]
\centering
\caption{Ablation study on reconstruction error formulation. Values are percentages.}
\label{tab:ablation_error}
\begingroup
\setlength{\tabcolsep}{5pt}
\renewcommand{\arraystretch}{1.08}
\scalebox{0.75}{\begin{tabular}{l l c c c}
\toprule
\textbf{Error Type} & \textbf{Description} & \textbf{Acc.} & \textbf{F1} & \textbf{AUROC} \\
\midrule
XCLIP only & Only semantic tokens & 79.56 & 77.14 & 82.78 \\
Signed error & $x_t-\hat{x}_t$ & \best{88.33} & \best{88.18} & \best{95.15} \\
Absolute error & $|x_t-\hat{x}_t|$ & 62.33 & 69.34 & 90.04 \\
Squared error & $(x_t-\hat{x}_t)^2$ & 81.11 & 86.49 & 94.73 \\
\bottomrule
\end{tabular}}
\endgroup
\vspace{-6mm}
\end{table}
\vspace{-4mm}
\subsubsection{Effect of Temporal Modeling.}
Table~\ref{tab:ablation_temporal} compares different temporal aggregation modules. Mamba achieves the best Accuracy and F1-score, while mean pooling and Transformer~\cite{vaswani2017attention} perform worse despite higher AUROC or more parameters.

ResNet50~\cite{he2016resnet}, LSTM~\cite{hochreiter1997lstm}, and 1D CNN~\cite{bai2018tcn} also underperform Mamba, supporting the use of an efficient sequence module for local semantic-error trajectories.
\begin{table}[!htbp]
\centering
\caption{Ablation study on temporal modeling modules. Values are percentages.}
\label{tab:ablation_temporal}
\begingroup
\setlength{\tabcolsep}{5pt}
\renewcommand{\arraystretch}{1.08}
\scalebox{0.75}{\begin{tabular}{l c c c c}
\toprule
\textbf{Temporal Module} & \textbf{Acc.} & \textbf{F1} & \textbf{AUROC} & \textbf{Params} \\
\midrule
Mean pooling & 68.55 & 75.63 & \best{96.17} & 122.055M \\
Transformer~\cite{vaswani2017attention} & 68.11 & 75.25 & 91.23 & 147.313M \\
ResNet50~\cite{he2016resnet} & 61.22 & 70.16 & 92.76 & 147.653M \\
LSTM~\cite{hochreiter1997lstm} & 64.22 & 70.36 & 90.99 & 128.354M \\
1D CNN~\cite{bai2018tcn} & 67.22 & 74.18 & 91.63 & 128.350M \\
Mamba, ours & \best{88.33} & \best{88.18} & 95.15 & 129.097M \\
\bottomrule
\end{tabular}}
\endgroup
\vspace{-6mm}
\end{table}
Overall, the ablation results validate the complementary roles of WF-VAE reconstruction errors, semantic alignment, patch-level correspondence, and Mamba temporal modeling.
\vspace{-4mm}
\section{Conclusion}
\vspace{-4mm}
In this paper, we proposed ReConFuse, a reconstruction-guided semantic-error fusion framework for AI-generated video detection. By aligning WF-VAE reconstruction errors with XCLIP semantic tokens and modeling their temporal evolution, ReConFuse captures complementary spatial, semantic, and temporal evidence. Experiments on GenVideo and GenBuster demonstrate its effectiveness and generalization ability.
\bibliographystyle{splncs04}
\bibliography{references}
\end{document}